%% file: journal.tex
\documentclass[journal]{IEEEtran}

\usepackage{multicol}
\usepackage[utf8]{inputenc} 
\usepackage[T1]{fontenc}    
\usepackage{hyperref}       
\usepackage{url}            
\usepackage{booktabs}       
\usepackage{amsfonts}       
\usepackage{nicefrac}       
\usepackage{microtype}      
\usepackage{amsmath}
\usepackage{multirow}
\usepackage{graphicx}
\usepackage{comment}

\usepackage{color}
\usepackage{array}
\usepackage{bm}
\usepackage{bbm}

\input{include}

\usepackage{soul}

\usepackage[flushleft]{threeparttable}
\newcommand{\specialcell}[2][c]{%
  \begin{tabular}[#1]{@{}c@{}}#2\end{tabular}}
\newcolumntype{H}{>{\setbox0=\hbox\bgroup}c<{\egroup}@{}}
\title{PanRep: Graph neural networks for extracting universal node embeddings in heterogeneous graphs.}

\author{Vassilis~N.~Ioannidis,
        Da~Zheng,
         George~Karypis
\thanks{The work in this paper has been supported by Amazon Web Services (AWS).}
\thanks{V. N. Ioannidis, D. Zheng and G. Karypis are with the AWS Artificial Intelligence.}
}

\begin{document}

\maketitle

\begin{abstract}
Learning unsupervised node embeddings facilitates several downstream tasks such as node classification and link prediction. A node embedding is universal if it is designed to be used by and benefit various downstream tasks. This work introduces PanRep, a graph neural network (GNN) model, for unsupervised learning of universal node representations for heterogenous graphs. PanRep consists of a GNN encoder that obtains node embeddings and four decoders, each capturing different topological and node feature properties. Abiding to these properties the novel unsupervised framework learns universal embeddings applicable to different downstream tasks.
PanRep can be furthered fine-tuned to account for possible limited labels. In this operational setting PanRep is considered as a pretrained model for extracting node embeddings of heterogenous graph data. PanRep outperforms all unsupervised and certain semi-supervised methods in node classification and link prediction, especially when the labeled data for the semi-supervised methods is small. PanRep-FT (with fine-tuning) outperforms all other semi-supervised approaches, which corroborates the merits of pretraining models. Finally, we apply PanRep-FT for discovering novel drugs for Covid-19. We showcase the advantage of universal embeddings in drug repurposing and identify several drugs used in clinical trials as possible drug candidates.
\end{abstract}

\section{Introduction}
Learning node representations from heterogeneous graph data powers the success of many downstream machine learning tasks such as node classification~\cite{kipf2016semi} and link prediction~\cite{wang2017knowledge}. Graph neural networks (GNNs) learn node embeddings by applying a sequence of nonlinear operations parametrized by the graph adjacency matrix and achieve state-of-the-art performance in the aforementioned downstream tasks. 
The era of big data provides an opportunity for machine learning methods to harness large datasets~\cite{wu2013data}. Nevertheless, typically the labels in these datasets are scarce due to either lack of information or increased labeling costs~\cite{bengio2012unsupervised}. The lack of labeled data points hinders the performance of supervised algorithms, which may not generalize well to unseen data and motivates \emph{unsupervised} learning. 


Unsupervised node embeddings may be used for downstream learning tasks, while the specific tasks are typically not known a~priori. For example, node representations of the Amazon book graph can be employed for recommending new books as well as classifying a book's genre. 
Unsupervised learning can also be employed when limited labels of the downstream task are available in two ways. First, the node embeddings learned in the unsupervised setting can be used as fixed features in a supervised learning method that will train with the limited labels. Such a two-stage approach will reduce the risk of overfitting since the node embeddings are not trained with the labels.
Second, refining the unsupervised node embeddings with these labels could further increase the representation power of the embeddings. This can be achieved by \emph{fine-tuning} the unsupervised model. In this setting, the GNN model is initialized with the trained model weights obtained by unsupervised learning. Next the model is extended by an output layer and trained for a few epochs with the available labels in an end-to-end fashion. Natural language processing methods have achieved state-of-the-art performance by applying such a fine-tuning  framework~\cite{devlin2018bert}.  Fine-tuning pretrained models is beneficial compared to end-to-end supervised learning since the former typically generalizes better especially when labeled data are limited and decreases the inference time since typically just a few fine-tuning iterations typically suffice for the model to converge~\cite{erhan2010does}. In this work we will study both applications of unsupervised learning with limited label information. 

%
This work aspires to provide \emph{universal} node embeddings, which will be applied in multiple downstream tasks and achieve comparable performance to their supervised counterparts. 
The paper puts forth a novel framework for unsupervised learning of universal node representations on heterogenous graphs termed PanRep\footnote{Pan: Pangkosmios (Greek for universal) and Rep: Representation}
. 
It consists of a GNN encoder that maps the heterogenous graph data to node embeddings and four decoders, each capturing different topological and node feature properties. The cluster and recover (CR) decoder exploits a clustering prior of the node attributes. The motif (Mot) decoder captures structural node properties that are encoded in the network motifs. The meta-path random walk (MRW) decoder promotes embedding similarity among nodes participating in a MRW and hence captures intermediate neighborhood structure. Finally, the heterogeneous information maximization (HIM) decoder aims at maximizing  the mutual information among node local and the global representations per node type. These decoders model general properties of the graph data related to node homophily~\cite{gleich2015pagerank,kloster2014heat} or node structural similarity~\cite{rossi2014role,donnat2018learning}.  PanRep is solely supervised by the decoders and has no knowledge of the labels of the downstream task. 

\begin{figure*}
    \centering
    \includegraphics[scale=0.45]{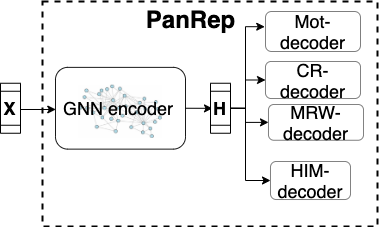}\hspace{1.2cm}\includegraphics[scale=0.45]{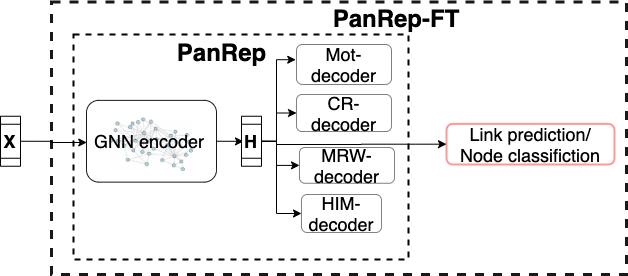}
    \caption{Illustration of the PanRep (left) and PanRep-FT (right) models. The GNN encoder processes the node features $\mathbf{X}$ to obtain the embeddings $\mathbf{H}$. The  decoders facilitate unsupervised learning of $\mathbf{H}$. 
    }
    \label{fig:panrep}
\end{figure*}

%
PanRep may utilize limited labels in two ways. First, the universal embeddings learned by PanRep are employed as features by models such as SVM~\cite{suykens1999least} or DistMult~\cite{yang2014embedding} to be trained for the downstream tasks. Such an approach will reduce the risk of overfitting since the universal embeddings encode general properties of the graph data.  
Second, this paper also considers  a fine-tuning PanRep (PanRep-FT) that inherits the weights of the PanRep model and employs an output layer for the label prediction. By selecting an appropriate output layer PanRep-FT accommodates arbitrary downstream tasks. In this operational setting, PanRep-FT is optimized adhering to a task-specific loss. PanRep can be considered as a pretrained model for extracting node embeddings of heterogenous graph data. Figure~\ref{fig:panrep} illustrates the two novel models.  

The contribution of this work is threefold.
\begin{itemize}
    \item[\textbf{C1.}] We introduce a novel problem formulation of universal unsupervised learning and design a tailored learning framework termed PanRep. We identify the following  general properties of the heterogenous graph data: (i) the clustering of local node features, (ii) structural similarity among nodes, (iii)  the local and intermediate neighborhood structure, (iv) and the mutual information among same-type nodes. We develop four tasks specific decoders to model the aforementioned properties. 
    
\item[\textbf{C2.}] We extend the unsupervised universal learning framework to directly account for labels of the downstream task. By appropriately extending PanRep's architecture we can model arbitrary downstream learning tasks like node classification or link prediction. PanRep-FT  refines the universal embeddings and  increases the model generalization capability.
%
    
\item[\textbf{C3.}]We compare the proposed models to state-of-the-art semi-supervised and unsupervised methods for node classification and link prediction. PanRep outperforms all unsupervised and certain semi-supervised methods in node classification, especially when the labeled data for the semi-supervised methods is small. PanRep-FT outperforms even semi-supervised approaches in node classification and link prediction, which corroborates the merits of pretraining models. Finally, we apply our method on the drug-repurposing knowledge graph (DRKG) for discovering drugs for Covid-19 and identify  several drugs used in clinical trials as possible drug candidates.
\end{itemize}   

The rest of the paper is structured as follows. 
Sec.~\ref{sec:rel} introduces some related works to PanRep.
Sec.~\ref{sec:prob} reviews some preliminaries to introduce the main paper. Sec.~\ref{sec:panrep} introduces the proposed PanRep model.
The experimental setup is detailed in Sec.~\ref{sec:expsetup}.
Finally, results and conclusions are presented in Secs.~\ref{sec:res} and~\ref{sec:cl}, respectively. 
\section{Related work}
\label{sec:rel}
Deep learning architectures typically process the input information using a succession of $L$ hidden layers. Each of the layers comprises a conveniently parametrized linear transformation, a scalar nonlinear transformation, and possibly a dimensionality reduction (pooling) operator. By successively combining (non)linearly local features, the aim at a high level is to progressively extract useful information for learning~\cite{goodfellow2016deep,gama2020graphs}. GNNs tailor these operations to the graph that supports the data~\cite{bronstein2017geometric}, including the linear \cite{defferrard2016convolutional}, and nonlinear \cite{defferrard2016convolutional,gama2018convolutionaljournal,ioanniditensorgcnstspjournal,ioannidis2018graphrnn,luana2018median,ioannidis2020efficient} operators. PanRep is an unsupervised GNN model that operates even in the absence of labeled data, whereas PanRep-FT provides a finetuned version of PanRep for supervised learning.  The following sections introduce three research areas that this paper also explores with introducing relevant related work.

\subsection{Unsupervised learning}  Representation learning amounts to mapping nodes in an embedding space where the graph topological information and structure is preserved~\cite{hamilton2017representation}. Typically, representation learning methods follow the encoder-decoder framework advocated by PanRep. Nevertheless, the decoder is typically attuned to a single task based on e.g., matrix factorization~\cite{tang2015line,ahmed2013distributed,cao2015grarep,ou2016asymmetric}, random walks~\cite{grover2016node2vec,perozzi2014deepwalk}, or kernels on graphs~\cite{smola2003kernels}. Recently, methods relying on GNNs are increasingly popular for representation learning  tasks~\cite{wu2020comprehensive}. GNNs typically rely on random walk-based objectives~\cite{grover2016node2vec,hamilton2017representation} or on maximizing the mutual information among node representations~\cite{velivckovic2018deep}. Relational GNNs methods extend representation learning to heterogeneous graphs~\cite{dong2017metapath2vec,shi2018heterogeneous,shang2016meta}. Relative to these contemporary works PanRep introduces multiple decoders to learn universal embeddings for heterogeneous graph data capturing the clustering of local node features, structural similarity among nodes,   the local and intermediate neighborhood structure, and the mutual information among same-type nodes. 

\subsection{Supervised learning} Node classification is typically formulated as a semi-supervised learning (SSL) task over graphs, where the labels for a subset of nodes are available for training~\cite{belkin2004regularization}. 
Graph-based SSL methods typically assume that the true labels are smooth over the graph,  which naturally motivates leveraging the network topology to propagate the labels and increase learning performance.  Graph-induced smoothness can be captured by graph kernels~\cite{belkin2006manifold,smola2003kernels,ioannidis2018kernellearn};  Gaussian random fields \cite{zhu2003semi}; or  low-rank {parametric} models~\cite{shuman2013emerging,ceci2020graph,ceci2020graphunc,segarra2015percolation,ioannidis2018coupledcom}. 
GNNs achieve state-of-the-art performance in SSL by utilizing regular graph convolution~\cite{kipf2016semi} or graph attention~\cite{velivckovic2017graph}, while these models have been extended in heterogeneous graphs~\cite{schlichtkrull2018modeling,fu2020magnn,wang2019heterogeneous}. 

Similarly, another prominent supervised downstream learning task is link prediction with numerous applications in recommendation systems~\cite{wang2017knowledge} and drug discovery~\cite{zhou2020network,drkg2020}.  Knowledge-graph (KG) embedding models rely on mapping the nodes and edges of the KG to a vector space by maximizing a score function for existing KG edges~\cite{wang2017knowledge,yang2014embedding,zheng2020dgl}.  RGCN models~\cite{schlichtkrull2018modeling} have been successful in link prediction and contrary to KG embedding models can further utilize node features. The universal embeddings extracted from PanRep without labeled supervision offer a strong competitive to these supervised approaches for both node classification and link prediction tasks.

\subsection{Pretraining} Pretraining models provides  a significant performance boost compared to traditional approaches in natural language processing~\cite{devlin2018bert} and computer vision~\cite{donahue2014decaf,girshick2014rich} . Pretraining offers increased generalization capability especially when the labeled data is scarce and increased inference speed relative to end-to-end training~\cite{devlin2018bert}. Parallel to our work \cite{hu2020gpt,qiu2020gcc} proposed pretraining models for GNNs.  These contemporary works supervise the models only using attribute or edge generation schemes without accounting for higher order structural similarity or maximizing the mutual information of the embeddings. Further, the work in~\cite{hu2020gpt} focuses solely on ranking objectives for evaluation. Recently,~\cite{hu2019strategies} introduced a framework for pretraining GNNs for graph classification.
Different than~\cite{hu2019strategies} that focuses on graph representations, PanRep aims at node prediction tasks and obtains node representations via capturing properties related to node homophily~\cite{gleich2015pagerank} or node structural similarity~\cite{rossi2014role}.
PanRep is a novel pretrained model for node classification and link prediction that requires significantly less labeled points to reach the performance of its fully supervised counterparts.

\section{Definitions and notation}
\label{sec:prob}
Scalars are denoted by lowercase, column vectors by bold lowercase, and matrices by bold uppercase letters. Superscript $\cdot^\top$ 
denotes the transpose of a matrix; and $\mbox{diag} ({\mathbf x})$ corresponds to a diagonal
matrix with the entries of $\mathbf 
x$ 	on its diagonal. 


A heterogeneous graph with $T$ node types and $R$ relation types is defined as $\mathcal{G}:=\{\{\mathcal{V}_{t}\}_{t=1}^{T},\{\mathcal{E}_r\}_{r=1}^{R}\}$\footnote{This paper considers unweighted graphs that is the weight in the edge is either $0$ or $1$.}. The node types represent the 
different entities and the relation types represent how these entities are semantically associated to each other. For example, in the IMDB network, the node types correspond to actors, directors, movies, etc., whereas the relation types correspond to \emph{directed-by} and \emph{played-in} relations. The number of nodes of type $t$ is denoted by $N_t$ and its associated nodal set by  $\mathcal{V}_{t}:=\{n_t\}_{n=1}^{N_t}$. The total number of nodes in $\mathcal{G}$ is $N:=\sum_{t=1}^t{N_t}$. The $r$th relation type, $\mathcal{E}_{r}:=\{(n_{t}, r ,n'_{t'})\in\mathcal{V}_{t}\times \mathcal{V}_{t'} \}$,  holds all interactions of a certain type among $\mathcal{V}_{t}$ and $\mathcal{V}_{t'}$ and may represent that a movie is \emph{directed-by} a director.  Heterogenous graphs are typically used to represent KGs~\cite{wang2017knowledge}. 

Each node $n_t$ is also associated with an $F\times1$ feature vector $\mathbf{x}_{n_t}$. This feature may be a natural language embedding of the title of a movie. The nodal features are collected in a $N\times F$ matrix $\mathbf{X}$. Note that certain node types may not have features and for these we use an embedding layer to represent their features. 




\section{PanRep}
\label{sec:panrep}
Given $\mathcal{G}$ and $\mathbf{X}$, our goal is to learn a function $g$ such that $\mathbf{H}:=g(\mathbf{X},\mathcal{G})$, where  $\mathbf{H}\in \mathbb{R}^{N\times D}$ represents the node embeddings and $D$ is the size of the embedding space. 
 Note that in estimating $g$, no labeled information is available.
 
Our work aims at universal representations $\mathbf{H}$ that perform well on different downstream tasks. Different node classification and link prediction tasks may arise by considering different number of training nodes and links and different label types, e.g., occupation label or education level label.  Consider $I$ downstream task, for the universal representations $\mathbf{H}$ it holds that
\begin{align}
    \mathcal{L}^{(i)}(f^{(i)}(\mathbf{H}),\mathcal{T}^{(i)})\le \epsilon,~i=1,\ldots,I,
\end{align}
where $\mathcal{L}^{(i)}$, ${f}^{(i)}$, and $\mathcal{T}^{(i)}$ represent the loss function, learned classifier, and training set (node labels or links) for task $i$, respectively and $\epsilon$ is the largest error for all tasks. The goal of unsupervised universal representation learning is to learn $\mathbf{H}$ such that  $\epsilon$ is small. While learning $\mathbf{H}$, PanRep  does not have knowledge of $\{\mathcal{L}^{(i)}, {f}^{(i)}, \mathcal{T}^{(i)}\}_i$. Nevertheless, by utilizing the novel decoder scheme PanRep achieves superior performance even compared to supervised approaches across tasks.

Our universal representation learning framework aims at embedding nodes in a low-dimensional space such that the representations are discriminative for node classification and link prediction. PanRep utilizes an GNN encoder that maps the node features and graph structure to node embeddings $\mathbf{H}$ that is detailed in Sec.~\ref{sec:encoder}. The node embedding matrix $\mathbf{H}$ is the input to four decoders each capturing unique graph properties; see Sec.~\ref{sec:univ}. The combined loss of the decoders trains PanRep in an end-to-end fashion. A task specific loss function for node classification and link prediction extends PanRep to supervised learning settings and gives rise to PanRep-FT in Sec.~\ref{sec:panrepft}. Figure~\ref{fig:panrep} provides an illustration of the overall PanRep architecture. Different than existing approaches, PanRep combines multiple decoders in a multi-task learning scheme to learn node embedding that perform well in a variety of tasks.


\subsection{PanRep Encoder}
\label{sec:encoder}
The abundance of graph-abiding data calls for advanced learning techniques that complement nicely standard machine learning tools when the latter cannot be directly employed, e.g. due to irregular data inter-dependencies. Permeating the benefits of deep learning to graph data, graph neural networks (GNNs) offer a versatile and powerful framework to learn from complex graph data~\cite{bronstein2017geometric}. 

The GNN layer operates per node $n$ in the graph and performs 3 steps: 1) aggregates the input embeddings corresponding to the neighbors of $n$, 2) Projects the combined embedding to a new vector via a parametrized projection matrix $\mathbf{W}$, and 3) Nonlinearly transforms the projected embedding.  Different GNN models employ variations of these steps.
Although the PanRep framework can utilize any GNN model as an encoder~\cite{wu2020comprehensive}, in this paper PanRep uses a relational graph convolutional network (RGCN) encoder~\cite{schlichtkrull2018modeling}. RGCNs extend the graph convolution operation~\cite{kipf2016semi} to heterogenous graphs.  The $l$th RGCN layer computes the $n$th node representation $\mathbf{h}^{(l+1)}_{n}$ as follows
\begin{align}
\mathbf{h}^{(l+1)}_{n}
	:=
	\sigma\left(\sum_{r=1}^{R}\sum_{ n'\in\mathcal{N}_n^{r}} 
	{\mathbf{h}}_{n'}^{(l)} \mathbf{W}^{(l)}_{r}\right),
	\label{eq:sem}
\end{align}
where $\mathcal{N}_n^{r}$ is the neighborhood of node $n $ under relation $r$, $\sigma$ the rectified linear unit non linear function, and $\mathbf{W}^{(l)}_{r}$ is a learnable matrix associated with the $r$th relation.  Essentially, the output of the RGCN layer for node $n$ is a nonlinear combination of the hidden representations of neighboring nodes weighted based on the relation type. The node features are the input of the first layer in the model i.e., $\mathbf{h}^{(0)}_{n}=\mathbf{x}_n$, where $\mathbf{x}_n$ is the node feature for node $n$. The matrix $\mathbf{H}$ in this paper represents the embedding extracted in the final layer.

Recent studies~\cite{mpnn,graphnets} manage to unify different GNN variants into the {message passing paradigm}. The RGCN layer in~\ref{eq:sem} can be interpreted under this framework where each node $n$ collects the messages send by his neighbors ${\mathbf{h}}_{n'}^{(l)}$ and transforms them. By implementing the message passing paradigm the deep graph learning (DGL)\footnote{\url{https://www.dgl.ai/}} library provides highly optimized GNN models~\cite{wang2019deep}.

%


\subsection{Universal supervision decoders}
\label{sec:univ}

Methods for learning over graphs  typically rely on modeling homophily of nodes that postulates neighboring vertices to have similar attributes~\cite{smola2003kernels,yuan2014exploiting} or structural similarity among nodes~\cite{rossi2014role}, where vertices involved in similar graph structural patterns possess related attributes~\cite{donnat2018learning}. 

Motivated by these methods we identify related properties encoded in the graph data. Clustering nodes based on their attributes provides a strong signal for node homophily~\cite{koren2009matrix}. Network motifs reveal the local structure information for nodes in the graph~\cite{ahmed2017graphlet}. Metapaths encode the heterogeneous graph neighborhood and indicate the local connectivity~\cite{dong2017metapath2vec}. Finally, maximizing the mutual information among embeddings declusters node representations and provides further discriminative information~\cite{velivckovic2018deep}. 
%
%
Towards capturing the aforementioned properties, we develop four novel neural network based decoders.
PanRep's encoder computes the embedding matrix $\mathbf{H}$ that is fed to the decoders, which supervise the learning by promoting graph properties in $\mathbf{H}$. 

\subsubsection{Cluster and recover supervision} Node attributes may reveal interesting properties of nodes, such as clusters of customers based on their buying power and age. This is important in recommendation systems, where traditional matrix factorization approaches~\cite{koren2009matrix} rely on revealing clusters of similar buyers. 
To capitalize such information we propose to supervise the universal embeddings by  such cluster representations. Specifically, we cluster the node attributes via $K$-means~\cite{kanungo2002efficient} and then design a model that decodes $\mathbf{H}$ to recover the original clusters. The CR-decoder amounts to a single layer multilayer perceptron (MLP) as follows
\begin{align}
\hat{\mathbf{C}}:=\sigma(\mathbf{H}\mathbf{W}_{\textsc{cr}}),
\end{align}
where $\hat{\mathbf{C}}$ is a $N\times K$ matrix representing the output of the decoder and $\mathbf{W}_{\textsc{cr}}$ is a learnable matrix.
The task specific loss function for this decoder is 
\begin{align}
    \mathcal{L}_{\textsc{cr}}:=-\sum_{n=1}^{N}\sum_{k=1}^K 
C_{nk}\ln{\hat{C}_{nk}},
\label{eq:cr}
\end{align}
where the cluster assignment $C_{nk}$ is 1 if node $n$ belongs in class $k$ and the predicted cluster assignment $\hat{C}_{nk}$ is the output of the CR-decoder. We showcase in the experiments that this decoder is superior to the attribute generation scheme used in~\cite{hu2020gpt}. Such a supervision decoder will enrich the universal embeddings $\mathbf{H}$ with information based on the clustering of local node features. 

\subsubsection{Motif supervision} Network motifs are sub-graphs where the nodes have specific connectivity patterns. Typical size-3 motifs for example, are the triangle and the star motifs. Each of these sub-graphs conforms to a particular pattern of interactions among nodes, and reveals important properties for the participating nodes.  In gene regulatory networks for example, motifs relate to 
 certain biological properties~\cite{babu2004structure}.   The work in~\cite{ahmed2017graphlet} develops efficient parallel implementations for extracting network motifs. We aspire to capture structural similarity among nodes by predicting their motif information. The motivation  is that nodes which might be distant in the graph may have similar structural properties as described by their motifs. 

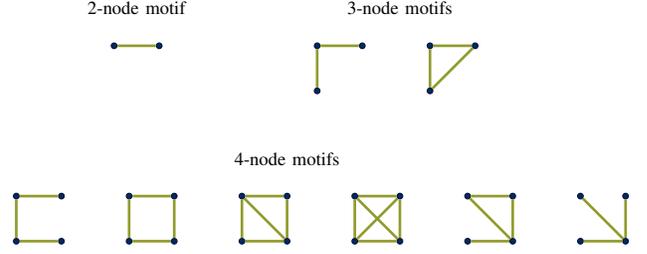
\begin{figure}
\centering{\input{motifs}}
\caption{Connected motifs up to size 4 nodes. The motif frequency $\bm{\mu}_n$ shows how many times node $n$ exits in the corresponding motifs.}
\label{fig:motifs}
\end{figure}

Using the method in~\cite{ahmed2017graphlet} we extract  a $M\times 1$ frequency vector $\bm{\mu}_n$ per node that shows how many times the node $n$ participates in motifs of size up to 4 nodes. Fig.~\ref{fig:motifs} shows the different motifs up to size 4.
%
%
This information reveals the structural role of nodes  such as star-center, star-edge nodes, or bridge nodes.
The motif decoder predicts this vector for all nodes using the universal representation $\mathbf{H}$. This allows for information sharing among nodes which are far away in the graph but have similar motif frequency vectors. 
The motif decoder utilizes a single layer MLP as follows
\begin{align}
\hat{\bm{\mu}}_n:=\sigma(\mathbf{h}_n\mathbf{W}_{\textsc{mot}}),~~ n=1,\ldots,N,
\end{align}
where $\hat{\bm{\mu}}_n$ is a $M\times 1$ matrix representing the output of the decoder and $\mathbf{W}_{\textsc{mot}}$ is a $D\times M$ learnable  matrix.
The motif decoder optimizes the following loss function
\begin{align}
    \mathcal{L}_{\textsc{mot}}:=\sum_{n=1}^N\|\bm{\mu}_n-\hat{\bm{\mu}}_n\|_2^2,
    \label{eq:mot}
\end{align}
where $\hat{\bm{\mu}}_n$ is the output of the Mot-decoder for the $n$th node. Using the Mot-decoder PanRep enhances the universal embeddings by structural information encoded in the node motifs.

\subsubsection{Metapath RW supervision} Metapaths are sequences of edges of possibly different type that connect nodes in a KG. A metapath random walk (MRW) is a specialized RW that follows different edge-types~\cite{dong2017metapath2vec}. 

We aspire to capture local connectivity patterns by promoting nodes participating in a MRW to have similar embeddings. Consider all node pairs for nodes $(n_t,{n'}_{t'})$  participating in a MRW, the following criterion maximizes the similarity among these nodes as follows
\begin{align}
    \mathcal{L}_{\textsc{mrw}}:= \log(1+\exp(-y\times \mathbf{h}_{n_t}^\top \mbox{diag}(\mathbf{r}_{t,t'})\mathbf{h}_{{n'}_{t'}})),\label{eq:mrw}
\end{align}
where $\mathbf{h}_{n_{t}}$ and $\mathbf{h}_{{n'}_{t'}}$ are the universal embeddings for nodes $n_t$ and ${n'}_{t'}$, respectively, $\mathbf{r}_{t,t'}$ is an embedding parametrized on the pair of node-types and $y$ is $1$ if $n_t$ and ${n'}_{t'}$ co-occur in the MRW and $-1$ otherwise. Negative examples are generated by randomly selecting tail nodes for a fixed head node with ratio 5 negatives per positive example. 

Metapaths convey more information than regular links since the former can be defined to promote certain prior knowledge. For example, in predicting the movie genre in IMDB the metapath configured by the edge types (played by, played in) among node types (movie, actor, movie) will potentially connect movies with same genre, which will allow the PanRep to map such movies to similar embeddings and  hence it is desirable. The embedding per node-type pair $\mathbf{r}_{t,t'}$ allows the MRW-decoder to weight the similarity among node embeddings from different node types accordingly. The length of the MRW controls the radius of the graph neighborhood in equation~\eqref{eq:mrw} and it can vary from local to intermediate. 
Note that link prediction is a special case of MRW supervision that considers MRWs of length $1$. 


\subsubsection{Heterogenous information maximization} The aforementioned supervision decoders capture clustering affinity, structural similarity and local and intermediate neighborhood of the nodes. Nevertheless, further information can be extracted by the representations by maximizing the mutual information among node representations. Such an approach for homogeneous graphs is detailed in~\cite{velivckovic2018deep}, where the  mutual information between node representations and the global graph summaries is maximized~\cite{hjelm2018learning}. 

Towards further refining the universal embeddings, we propose a generalization of~\cite{velivckovic2018deep} for heterogeneous graphs. We consider a global summary vector per $t$  as $\mathbf{s}_{t}:=\sum_{n_t=1}^{N_t}\mathbf{h}_{n_t}$ that captures the average $t$th node representation. We aspire to maximize the mutual information among $\mathbf{s}_{t}$ and the corresponding nodes in $\mathcal{V}_{t}$. The proposed HIM decoder optimizes the following contrastive loss function
\begin{align}
    \mathcal{L}_{\textsc{him}}:= \sum_{t=1}^T \bigg( \sum_{n_t=1}^{N_t}&\log\big(\sigma(\mathbf{h}_{n_t}^\top\mathbf{W}
    \mathbf{s}_{t})\big)\nonumber\\&+\log\big(1-\sigma(\tilde{\mathbf{h}}_{n_t}^\top\mathbf{W}
    \mathbf{s}_{t})\big)\bigg),
    \label{eq:him}
\end{align}
where the bilinear scoring function $\sigma(\tilde{\mathbf{h}}_{n_t}^\top\mathbf{W}
\mathbf{s}_{t})$ captures how close is $\mathbf{h}_{n_t}$ to the global summary, $\mathbf{W}$ is a learnable matrix and $\tilde{\mathbf{h}}_{n_t}$ represents the negative example used to facilitate training. Designing negative examples is a cornerstone property for training contrastive models~\cite{velivckovic2018deep}. We generate the negative examples in~\eqref{eq:him} by shuffling node attributes among nodes of the same type. The HIM decoder maximizes the mutual information across nodes and complements the former decoders.


\subsubsection{The overall loss function}
RGCN extracts the embedding matrix $\mathbf{H}$ that is the input to the decoders presented in~\ref{sec:univ}. PanRep's overall loss function is the linear combination of \eqref{eq:cr}, \eqref{eq:mot}, \eqref{eq:mrw}, and \eqref{eq:him} 
\begin{align}
    \mathcal{L}:=\mathcal{L}_{\textsc{cr}}+\mathcal{L}_{\textsc{mot}}+\mathcal{L}_{\textsc{mrw}}+\mathcal{L}_{\textsc{him}}.\label{eq:ov}
\end{align}
A backpropagation algorithm~\cite{rumelhart1986learning} minimizes \eqref{eq:ov}. 
The objective also relates to the framework of deep multitask learning~\cite{zhang2014facial}, since the GNN encoder is shared across the multiple supervision tasks and PanRep makes multiple inferences in one forward pass. Such networks are not only scalable, but the shared features within these networks can induce more robust regularization and possibly boost performance~\cite{chen2017gradnorm}. 
Introducing adaptive weights per decoder to control its learning rate is a future direction of PanRep.

\subsection{PanRep-FT}
\label{sec:panrepft}
In certain cases a subset of labels may be known a~priori for the downstream task and it is beneficial to fine-tune PanRep's model to obtain refined node representations. In this context, PanRep is a pretrained model and a downstream task specific loss can further supervise PanRep. First, we train PanRep for some epochs using the unsupervised decoders in Sec.~\ref{sec:univ} and then we train PanRep's encoder for some epochs solely on the downstream task specific loss. Fig.~\ref{fig:panrep} illustrates on the right side the PanRep-FT framework. Next, we present the supervised tasks along with their corresponding problem formulations and training losses. 

\subsubsection{Node classification}
Node classification has numerous applications  and gives rise a SSL task over graphs, where given the features $\mathbf{X}$, the graph structure $\mathcal{G}$,  and the labels for a subset of nodes the goal is to predict the labels across all nodes~\cite{belkin2004regularization}. 

Each node ${n}$ has a label $y_{n}\in\{0,\ldots,P-1\}$, which in the IMDB network may represent the genre of a movie. In SSL, we know labels only for a subset of nodes $\{y_{{n}}\}_{{n}\in\mathcal{M}}$, with $\mathcal{M} \subset\mathcal{V}$. This partial availability may be attributed to privacy concerns (medical data); energy considerations  (sensor networks); or unrated items (recommender systems). The ${N}\times P$ matrix  $\mathbf{Y}$ is the one-hot representation of the true nodal labels; that is, if $y_{n}=p$ then $Y_{{n},p}=1$ and $Y_{{n},p'}=0, \forall p'\ne p$.

Given the universal embeddings $\mathbf{H}$ PanRep-FT employs a single layer MLP to obtain the predicted label matrix $\hat{\mathbf{Y}}$  as follows 
\begin{align}
\hat{\mathbf{Y}}:=\mathbf{H}\mathbf{W}_\textsc{nc},
\label{eq:ncpred}
\end{align}
where the projection matrix $\mathbf{W}_\textsc{nc}$ is a learnable parameter,  $\hat{\mathbf{Y}}$ is an ${N}\times K$ matrix, and
$\hat{Y}_{{n},k}$ represents the probability that $y_{n}=k$.
PanRep-FT loss function for this node classification (nc) task is a cross-entropy loss
\begin{align}\mathcal{L}_{\textsc{nc}}:=-\sum_{{n}\in\mathcal{M}}\sum_{p=1}^P Y_{{n} p}\ln{\hat{Y}_{{n} p}},
\label{eq:ncloss}
\end{align}
where the error is averaged over the nodes with labels in $\mathcal{M}$. The loss function in~\eqref{eq:ncloss} will force PanRep's encoder to learn suitable parameters to minimize the classification error. After training the matrix $\hat{\mathbf{Y}}$  from the output function in~\eqref{eq:ncpred} predicts the class of the unlabeled nodes.

\subsubsection{Link prediction} 
Consider the heterogeneous graph  $\mathcal{G}$ in Sec.~\ref{sec:prob}.  Given the sets of links $\{\mathcal{E}_{r}\}_{r=1}^{R}$, and the node features $\mathbf{X}$ the goal of link prediction is to predict whether a set of links different than the one used for training might exist or not in the graph. 

Typically, link prediction models utilize a contrastive loss function that requires the model to distinguish among positive and negative examples~\cite{zheng2020dgl}. In this context, positive examples are the set of existing links in the graph. The negative examples, which are links that the model should classify as nonexistent, are typically sampled from the missing links in the graph. For each \emph{positive triplet} $q=(n_t,{r},{n'}_{t'})$ a number of negative links is generated by corrupting the head and tail entities at random $(n_t,{r},{\nu'}_{t'})$ and $(\nu_t,{r},{n'}_{t'})$.  This paper considers 5 negative examples per one positive for link prediction.

PanRep-FT employs a DistMult model~\cite{yang2014embedding} for link prediction. The loss function is
\begin{align}
\label{eq:rgcnsup}
\min \sum_{(n_t,{r},{n'}_{t'}) \in \mathbb{D}^+ \cup \mathbb{D}^-} \log(1+\exp(-y \times
\mathbf{h}_{n_t}^\top \mbox{diag}(\mathbf{h}_r)\mathbf{h}_{{n'}_{t'}})),
\end{align}
where $\mathbf{h}_{n_t}$ and $\mathbf{h}_{{n'}_{t'}}$ are the embedding of
the head and tail entity $n_t$, ${n'}_{t'}$ obtained by PanRep's encoder and   $\mathbf{h}_r$ is a trainable parameter corresponding to relation ${r}$ that is directly from~\eqref{eq:rgcnsup}. The scalar represented by $\mathbf{h}_{n_{t}}^{\top} \text{diag}{(\mathbf{h}_r)} \mathbf{h}_{{n'}_{t'}}$ denotes the score of triplet $(n_t,{r},{n'}_{t'})$ as given by the DistMult model~\cite{yang2014embedding}. Finally, $\mathbb{D}^+$ and $\mathbb{D}^-$ are the positive and negative sets of triplets and $y=1$ if the triplet corresponds to a positive example and $-1$ otherwise. By minimizing~\eqref{eq:rgcnsup} PanRep-FT will learn node embeddings $\mathbf{H}$, which will be informative for link prediction. PanRep-FT predicts new links by calculating their DistMult score; see also Sec.~\ref{sec:eval}.

Pretraining PanRep before applying the task specific loss functions may increase the model performance. 
BERT models in natural language processing have reported state of the art results by considering such a pretrain and fine-tune framework~\cite{devlin2018bert}. PanRep-FT is a counterpart of BERT for extracting information from heterogenous graph data. PanRep-FT combines the benefit of universal unsupervised learning and task specific information and achieves greater generalization capacity especially when labeled data are scarce~\cite{erhan2010does}. 
 

\textbf{Alternative encoders.} Several works consider designing possibly more general GNN encoders that utilize attention mechanism~\cite{wang2019heterogeneous,velivckovic2017graph} or graph isomorphism networks~\cite{xu2018powerful}. This paper proposes novel supervision decoders for unsupervised learning that capture general properties of the graph data. Designing a universal encoder based on these contemporary GNN models is a future direction of PanRep.
\section{Experimental setup}
\label{sec:expsetup}
We implement PanRep in the efficient deep graph learning (DGL)\footnote{\url{https://www.dgl.ai/}} library~\cite{wang2019deep}. 
PanRep experiments run on an AWS P3.8xlarge instances with 8 GPUs each having 16GB of memory\footnote{https://aws.amazon.com/ec2/instance-types/p3/}.  
\subsection{Methods}
Our universal represention learning techniques compares against with state-of-the-art methods. For node classification consider the following contemporary methods RGCN~\cite{schlichtkrull2018modeling},  HAN~\cite{wang2019heterogeneous}, MAGNN~\cite{fu2020magnn}, 
node2vec~\cite{grover2016node2vec}, 
meta2vec~\cite{dong2017metapath2vec} and an adaptations of the work in~\cite{velivckovic2018deep} for heterogenous graphs termed HIM. The competing methods RGCN, MAGNN and HAN also use the DGL. 
For link prediction the baseline models is RGCN~\cite{schlichtkrull2018modeling} with DistMult supervision~\cite{yang2014embedding} that uses the same encoder as PanRep. 

The parameters for all methods considered optimize the performance on the validation set. For PanRep the Mot-decoder and RC-decoder employ a 1-layer MLP. For the MRW-decoder we use length-$2$ MRWs. For the majority of the experiments PanRep uses a hidden dimension of 300, 1 hidden layer, 800 epochs of model training, 100 epochs for finetuning, and an ADAM optimzer~\cite{kingma2015adam} with  a learning rate of 0.001. For link prediction finetuning PanRep uses a DistMult model~\cite{yang2014embedding} whereas for node classification it uses a logistic loss.

\subsection{Datasets} 
We consider a subset of the IMDB dataset~\cite{IMDB} containing 11,616 nodes belonging to three node-types and 17,106 edges belonging to six edge-types. Each movie is associated with a label representing its genre and with a feature vector corresponding to its keywords. We also use a subset of the OAG dataset~\cite{OAG} with 23,696 nodes belonging to four node-types (authors, affiliations, papers, venues) and 90,183 edges belonging to 14 edge-types. In OAG we did not use $\textsc{mot}$ supervision since the graph does not have a rich motif structure. Each paper is associated with a label denoting the scientific area and with an embedding of the papers' text. Table~\ref{tab:dataset} provides additional information about these two datasets. 

\begin{table}[t]
\begin{center}
  \caption{Dataset statistics. The reverse edges are also present but do not appear in the table.}
  \label{tab:dataset}
  \scalebox{.95}{ 
  \begin{tabular}{lllll}
    \toprule
     Dataset & Node type & Nodes & Edge type & Edges \\
    \midrule
    IMDB & Movie (M) & 4,278 &  M-directed by-D & 4,278 \\& Director (D) & 2,081&M-played by-A & 12,828 \\ &Actor (A) & 5,25  &&   \\
    \midrule
    OAG & Author (A)& 13,720&A-writes as last-P& 4,522  \\ &Paper (P)& 7,326& P-in journal-V& 3,941 \\ &Affiliation (Af) & 2,290&P-conference-V&3,368 \\ &Venue (V)&782 &  P-cites-P& 3,327\\&&&A-writes as other-P& 7,769\\&&&A-writes as first-P& 4,795\\&&&A-affiliated with-Af&17,035\\
    \bottomrule
  \end{tabular}}
\end{center}
\end{table}

\begin{figure}%
    \centering
    \includegraphics[width=.90\columnwidth]{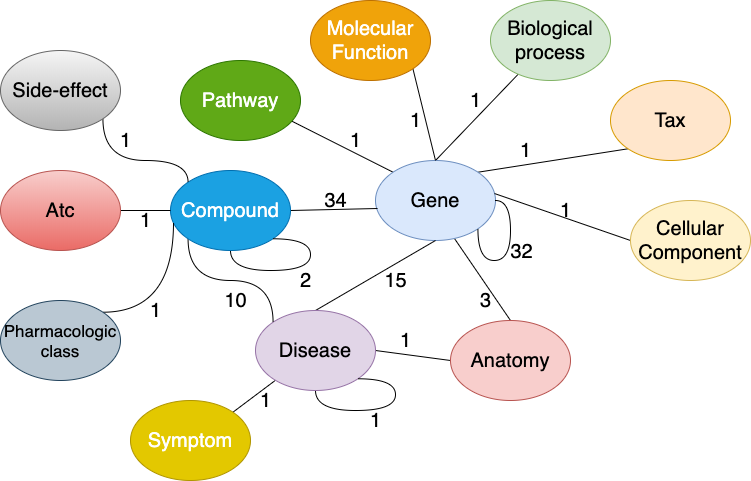}
    \caption{The schema of the Drug Repurposing Knowledge Graph (DRKG). The number next to an edge indicates the number of relation-types among the corresponding entity types in the DRKG.}
    \label{fig:schema}
\end{figure}

In addition, we use a third dataset corresponding to the drug repurposing knowledge graph (DRKG) constructed in~\cite{drkg2020}. DRKG, whose schema is shown in Figure~\ref{fig:schema}, contains 5,874,261 biological interactions belonging to 107 edge-types and 97,238 biological entities from 13 entity-types. DRKG is used for evaluating link-prediction tasks.

    
\subsection{Evaluation metrics}\label{sec:eval}
For evaluating node classification methods we use macro and micro F1 score that is advantageous when the classes are unbalanced~\cite{powers2011evaluation}. The F1 score is the harmonic mean of the precision and recall and reaches its best value at 1 and worst value at 0. The Mirco-F1 score calculates the metric across labels by counting the total true positives, false negatives and false positives. The Macro-F1 calculate metrics for each label, and find their unweighted mean. This does not take label imbalance into account.

Link prediction evaluation utilizes two standard criteria the Hit-10 and Mean Reciprocal Rank (MRR) metrics~\cite{bordes2013translating,poole2018simple,zheng2020dgl}.  Both metrics are derived by comparing how the score of the positive triplet relates to the scores of its associated negative instances. 

For each \emph{positive triplet} $q=(n_t,{r},{n'}_{t'})$ we generate all possible triplets of the form $(n_t,{r},{\nu'}_{t'})$ and $(\nu_t,{r},{n'}_{t'})$ by corrupting the head and tail entities. We then removed from them any triplets that already exist in the dataset. The set of triplets that remained form the \emph{negative triplets} associated with the initial positive triplet. The positive and negative triplets are then scored by the DistMult~\cite{yang2014embedding} model in~\eqref{eq:rgcnsup}.

We construct a list of triples $S_q$ containing $q$ and its associated negative triplets that order them in a non-increasing score fashion, and let $rank_q$ be $q$th's position in $S_q$. Given that,  Hit-10 is the average number of times the positive triplet is among the 10 highest ranked triplets,  whereas MRR is the average reciprocal rank of the positive instances.
Mathematically, the metrics are as follows
$$\mbox{Hit-}10:=\frac{1}{Q}\sum_{q=1}^Q \mathbbm{1}_{rank_q \le 10},~~~~\\
\mbox{MRR}:=\frac{1}{Q} \sum_{q=1}^Q \frac{1}{rank_q},$$
where $Q$ is the total number of positive triplets and $\mathbbm{1}_{rank_q\le 10}$ is 1 if $rank_q \le 10$, otherwise it is 0. Note that Hit-$10$ and MRR are between 0 and 1, with values closer to 1 corresponding to better predictions.

\subsection{Scalability} PanRep is implemented using DGL's mini-batch training, which is scalable w.r.t. time and space. During mini-batch training a subset of seed nodes is sampled along with the $K$-hop neighbors of the seed nodes, where $K$ is the depth of the GNN. Furthermore, the number of edges at each level is bounded by using neighbor sampling. This allows training on graphs with billions of nodes/edges since only a fraction of the nodes/edges will be loaded on the GPU per mini-batch. PanRep's complexity is controlled by the RGCN encoder whose computation complexity of a mini-batch is $\mathcal{O}(B F^K )$ for sparse graphs, where $B$ is the batch size, $K$ is the number of layers and $F$ is the maximum number of neighbors per node.  By using neighbor sampling, $F$ is usually less than 10--20 and in our experiments we found that K=2 often works well. The decoders' complexity is significantly lower and do not add but a constant factor to the overal complexity  and hence the overall per-epoch runtime is similar to that of RGCN.
\section{Results}
\label{sec:res}

\subsection{Node classification}
\label{sec:nodecl}
In this experiment we compare semi-supervised and unsupervised methods for classification.
First, the labeled nodes are split in 10\%  training, 5\% validation, and 85\% testing sets. The semi-supervised methods utilize the training labels whereas the unsupervised methods do not use them in calculating the node embeddings. 
Then the embeddings corresponding to the 85\% testing nodes are further split to training and testing sets and a linear support vector machine (SVM) is trained on the training set for evaluation following the setup in~\cite{fu2020magnn}. 
%
%
The reason that we selected a simple linear classifier as the SVM over a potentially more powerful non-linear one, e.g., an multi-layer perception (MLP) classifier, is because an SVM classifier allows us to directly compare the representation power of the different approaches, whereas an MLP classifier that employs a neural network will enhance the representation power of certain methods and result to unfair comparisons. 

\begin{table*}[tb]
\caption{Node classification results.}
\label{tab:node_class}\centering
 \footnotesize
{  
\begin{tabular}{|c|c|c|H c|H c| c| c|H H c|c|c|c|c|}
\hline
\multirow{2}{*}  & \multirow{2}{*}{Train \%} && \multicolumn{6}{c|}{Unsupervised}                          & \multicolumn{6}{c|}{Semi-supervised}   \\ \cline{3-15}
&                          & & LINE  & node2vec & ESim  & meta2vec  & HIM & PanRep & GCN   & GAT   & HAN   & MAGNN  & RGCN  & PanRep-FT           \\ \hline
\multirow{6}{*}{IMDB}&\multirow{3}{*}{Macro-F1}
&  40\%       & 45.45 & 50.63    & 50.09 & 47.57        & 55.21& \textbf{56.04}& 53.67 & 55.50 & 56.15 & \textbf{60.27}&58.48 &{59.49} \\ \cline{3-15}
&& 60\%       & 47.09 & 51.65    & 51.45 & 48.17        & 57.66 & \textbf{58.51} & 54.24 & 56.46 & 57.29 & \textbf{60.66}&58.42 &{59.86} \\ \cline{3-15}
&& 80\%       & 47.49 & 51.49    & 51.37 & 49.99        & 57.89 & \textbf{60.23}& 54.77 & 57.43 & 58.51 & {61.44}& 58.76&\textbf{61.49} \\ \cline{3-15}
&\multirow{3}{*}{Micro-F1}
& 40\%       & 46.92 & 51.77    & 51.21 & 48.17        &  55.11&\textbf{55.92} & {53.76} & 55.56 & 57.32 & \textbf{60.50}&58.64 &59.67 \\ \cline{3-15}
&& 60\%       & 48.35 & 52.79    & 52.53 & 49.87        & 56.57&\textbf{58.41} & {54.23} & 56.47 & 58.42 & \textbf{60.88} & 58.55&59.75\\ \cline{3-15}
&& 80\%       & 48.98 & 52.72    & 52.54 & 50.50        & 57.79 &\textbf{60.14} & {54.63} & 57.40 & 59.24 & {61.53} &58.89 &\bf 61.59\\ 
\hline
\multirow{6}{*}{OAG}&\multirow{3}{*}{Macro-F1}
&  40\%       & 62.37 &56.37   & 50.09 & \bf 65.75      & 50.54& {57.76}& 53.67 & 55.50 & 63.99 & 63.31& 64.68&\bf 64.72\\ \cline{3-15}
&& 60\%       & 63.01 & 57.01    & 51.45 & \bf 66.09       &51.98 & {59.72} & 54.24 & 56.46 & 64.25 & 63.42& 65.96&\bf 66.99 \\ \cline{3-15}
&& 80\%       & 64.05 &58.05  & 51.37 & \bf 65.75        & 53.25& {63.03}& 54.77 & 57.43 & 64.37 & 63.89& 67.67 &\bf67.90 \\ \cline{3-15}
&\multirow{3}{*}{Micro-F1}
& 40\%       & 70.17 &70.17    & 51.21 & 74.54       &71.91&\textbf{75.50} & {53.76} & 55.56 & 73.95 & 72.74&\bf 81.92&80.36 \\ \cline{3-15}
&& 60\%       & 70.95 & 70.95   & 52.53 & 74.96        & 73.89&\textbf{77.39} & {54.23} & 56.47 & 75.32 & 72.75 & 81.39 &\bf81.78\\ \cline{3-15}
&& 80\%       & 72.24 & 72.24    & 52.54 & 74.73        &75.31&\textbf{79.76} & {54.63} & 57.40 & 75.24 & 73.43 & 82.38& \bf83.17\\ 
\hline
\end{tabular}}
\end{table*}


\begin{table*}[tb]
\caption{Node classification results for different labeled supervision splits.  }
\label{tab:node_class_dif_splits}\centering
 \footnotesize
{  
\begin{tabular}{|c|c|c|c|c|c|c|c|c|c|c|c|c|c|c|c|}
\hline
\multirow{2}{*}{Datasets} &\multirow{2}{*}{Metrics}   & \multicolumn{2}{l|}{Train embeddings \%}  &\multicolumn{2}{c|}{5\%} &
 \multicolumn{2}{c|}{ 10\%}  &
 \multicolumn{2}{c|}{ 20\%}  
 \\ \cline{3-10}
 &  &    {Train\%}  & PanRep  & RGCN  & PanRep-FT  & RGCN  & PanRep-FT   & RGCN & PanRep-FT
   \\ \hline
 \multirow{6}{*}{IMDB}  

&\multirow{3}{*}{Macro-F1}& 40\%       & {56.04} & 55.12    & \bf56.85 &58.48 &\bf{59.49}&  61.30        & \bf63.14\\ \cline{3-10}%
& & 60\%       & {58.51} & 55.20    & \bf59.39&58.42 &\bf{59.86} & 60.98        &  \bf62.91
\\ \cline{3-10}
&& 80\%       & {60.23} & 55.55    & \bf61.27 &58.76&\textbf{61.49}  & 61.10        & \bf62.72\\\cline{2-10}

&\multirow{3}{*}{Micro-F1} & 40\%       &{55.92} & 55.27   &\bf 56.92&58.64 &\bf59.67 & 61.49       & \bf63.17
\\ \cline{3-10}
& & 60\%       & {58.41} & 55.39    &\bf 59.45& 58.55&\bf59.75 & 61.17        & \bf62.89
\\ \cline{3-10}
&& 80\%       & {60.14}&55.62    & \bf61.32&58.89 &\bf61.39 & 61.30      &\bf 62.75
\\ \cline{1-10}
 \multirow{6}{*}{OAG} &\multirow{3}{*}{Macro-F1}& 40\% &57.76&55.51 &\bf64.99  & 64.68&\bf 64.72& \bf67.07&65.31\\ \cline{3-10}
 &&60\%&59.72&55.99 &\bf66.62  & 65.96&\bf 66.99&\bf67.58 &66.25\\ \cline{3-10}
&& 80\%        &63.03& 56.36&\bf68.94  & 66.10&\bf68.60 &67.67 &\bf67.90\\ \cline{2-10}
&\multirow{3}{*}{Micro-F1}& 40\%  
&75.50&78.00 &\bf80.19  &\bf 81.92&80.36 &\bf82.57 &81.17\\ \cline{3-10}
&& 60\% &77.39&78.07 & \bf81.36 &81.39 &\bf81.78 & \bf81.74&81.34\\ \cline{3-10}
&& 80\% & 79.76&78.44 & \bf82.52 & 82.38& \bf83.17& 82.20&\bf82.31
\\ \hline
\end{tabular}}
\end{table*}

We report the Macro and Micro F1 accuracy for different training percentages of the 85\% nodes fed to the SVM classifier in Table \ref{tab:node_class}.
First, PanRep outperforms other unsupervised approaches as well as some semi-supervised approaches. In the $80\%$ splits,  PanRep outperforms even its semi-supervised counterpart RGCN that uses node labels for supervision. Metapath2vec~\cite{dong2017metapath2vec} reports competitive performance for OAG in Macro-F1 score but underperforms in Micro-F1. Nevertheless, in this experiment Metapath2vec only uses the best performing metapath that is  \emph{paper-venue-paper}, which is considerably better than that of most other metapaths. This way of selecting the metapath, gives Meta2vec an (unfair) advantage over PanRep, which uses all metapaths of length~2 as it computes a universal representation (along with the other supervision decoders). 
PanRep-FT  outperforms  RGCN that uses the same encoder, which is a testament to the power of pretraining models. Finally, PanRep-FT  matches and outperforms in certain splits the state-of-the-art MAGNN that uses a more expressive encoder. PanRep's universal decoders enhance the embeddings with additional discriminative power that results to improved performance in the downstream tasks. 

Table~\ref{tab:node_class_dif_splits} reports the accuracy of the PanRep-FT  and the encoder RGCN, which is trained directly for the semi-supervised learning task to obtain the embeddings. 
PanRep-FT consistently outperforms RGCN across most SVM splits, whereas PanRep-FT's advantage over RGCN decreases as more training data for the embeddings are available; see e.g., the last column with $20\%$ training, which is justifiable since more labels diminish the advantage of pretraining.\footnote{PanRep-FT performs only 100 finetuning epochs with labeled supervision, which is significantly less to the 800 epochs of labeled supervision by RGCN.}  Moreover even without labeled supervision the unsupervised embeddings of PanRep outperform the semi-supervised RGCN embeddings for 5\% training labels. This demonstrates the importance of using PanRep as a pretraining method. Finally, RGCN reports similar performance across SVM training splits for all columns, whereas PanRep-FT increases with more supervision. These results suggest that PanRep-FT's embeddings have higher generalization capacity.

\begin{table*}[t!]
\caption{Ablation study for different supervision decoders. }
\label{tab:node_clasabls}\centering
 \footnotesize
{  
\begin{tabular}{|c|c|c|c|c|c|c|c|c|}
\hline
{Datasets} &{Metrics}     & {Train \%} &  \textsc{him}  & \textsc{mrw}& \textsc{him}+\textsc{mrw} & \textsc{mot} & \textsc{cr} & PanRep        \\ \hline
\multirow{6}{*}{IMDB}&\multirow{3}{*}{Macro-F1} 
& 40\%       & 55.21 & 54.54 & 55.32   & 42.28&32.21      & \textbf{56.04} \\ \cline{3-9}
&& 60\%       & 57.66 & 56.12  & 57.24  & 43.41 &  35.16       &  \textbf{58.51}  \\ \cline{3-9}
&& 80\%       & 57.89 & 56.64  & 57.74   & 44.31 & 35.66       &  \textbf{60.23}\\ \cline{2-9}
&\multirow{3}{*}{Micro-F1}& 40\%       & 55.11 & 54.36& 55.53    & 43.66 & 39.13       &\textbf{55.92} \\ \cline{3-9}
&& 60\%       & 56.57 &55.91 & 56.25  &44.89 & 40.03        &\textbf{58.41}\\ \cline{3-9}
&& 80\%       & 57.79 &56.49 & 58.42   & 45.65 & 40.66        &\textbf{60.14}\\ \cline{1-9}
\multirow{6}{*}{OAG} &\multirow{3}{*}{Macro-F1}& 40\%&50.54 &55.92 &56.11 &- & 15.46&\bf57.76\\\cline{3-9}
 &&60\%&51.98 &58.40 &58.91 & -& 15.48&\bf59.72\\ \cline{3-9}
&& 80\%& 53.25&60.61 &61.74 &-   &     15.54 &\bf63.03\\ \cline{2-9}
&\multirow{3}{*}{Micro-F1}& 40\%  
&71.91& 74.39&74.65 & -&63.06& \bf75.50 \\ \cline{3-9}
&& 60\%&73.89 &76.76&76.33 &- &63.14  &\bf77.39\\ \cline{3-9}
&& 80\% &75.31&78.90&78.71 &- &63.59 &  \bf79.76 \\\hline
\end{tabular}}
\end{table*}

\begin{table}
\begin{center}
\begin{threeparttable}
    \caption{Drug inhibits gene scores for Covid-19.}
    \label{tab:predicted_drugs_gene}
     \scriptsize
    \begin{tabular}{@{\hspace{10pt}}ll@{\hspace{20pt}}ll@{\hspace{10pt}}}\toprule
    \multicolumn{2}{c}{\bf PanRep-FT}& \multicolumn{2}{c}{\bf RGCN}\\
         Drug name& \# hits &Drug name& \# hits\\\midrule
	Losartan&	232& Chloroquine &	69\\
	Chloroquine&	198&Colchicine&	41\\
	Deferoxamine&	104	&Tetrandrine&	40\\
	Ribavirin	&101&Oseltamivir&	37\\
	Methylprednisolone&	44&Azithromycin&	36\\
	Thalidomide	&41&	Tofacitinib	&33\\
	Hydroxychloroquine&	19&		Ribavirin&	32\\
	Tetrandrine&	13&Methylprednisolone&	30\\
	Eculizumab&	10&	Deferoxamine&	30\\
	Tocilizumab	&9&	Thalidomide&	25\\
	Dexamethasone&	7&Dexamethasone&	24	\\
	Azithromycin&	6	&Bevacizumab&	21\\
	Nivolumab&	5&	Hydroxychloroquine&	19\\
	Piclidenoson&	5&	Losartan&	19\\
	Oseltamivir&	5&		Ruxolitinib&	13\\
	& 	&Eculizumab&12\\
	&	&Sarilumab	&8\\
   	&&	Baricitinib&	6\\
         \bottomrule
    \end{tabular}
\end{threeparttable}
\end{center}    
\end{table}
 
\begin{table*}[t!]
\caption{Attribute generation supervision against clustering and recover supervision for different number of clusters $K$. }
\label{tab:node_clasclustvsreconstr}\centering
 \footnotesize
{  
\begin{tabular}{|c|c|c|c|c|c|c|c|}
\hline
{Datasets} &{Metrics}     & {Train \%} &  \textsc{Attribute generation}  & \textsc{cr} $K=2$&  \textsc{cr} $K=4$ &  \textsc{cr} $K=5$ &  \textsc{cr} $K=10$   \\ \hline
\multirow{6}{*}{IMDB}&\multirow{3}{*}{Macro-F1} 
& 40\%       & 17.52 &29.52 & 31.94   & \bf 32.21& 31.22      \\ \cline{3-8}
&& 60\%       & 17.53 & 30.75  & 34.50  &\bf 35.16 & 34.19          \\ \cline{3-8}
&& 80\%       & 17.30 & 3145  &35.02  & \bf 35.66 & 34.82       \\ \cline{2-8}
&\multirow{3}{*}{Micro-F1}
& 40\%       & 35.66 & 38.51& 39.40    &\bf 39.13 & 38.38        \\ \cline{3-8}
&& 60\%       & 35.70 &38.83 & 41.02  &\bf41.13 &40.03      \\ \cline{3-8}
&& 80\%       & 35.07 &39.54 & 41.16   & \bf41.39 & 40.66       \\ \cline{1-8}
\end{tabular}}
\end{table*}

\subsection{Link prediction} 
Our universal embedding framework is further evaluated for link prediction using the IMDB and OAG datasets. The MRW decoder is used to evaluate the performance of PanRep in link prediction. Specifically, the MRW score calculated by the scaled inner product $\mathbf{h}_{n_t}^\top \mbox{diag}(\mathbf{r}_{t,t'})\mathbf{h}_{{n'}_{t'}}$ in~\eqref{eq:mrw} is used to predict whether there exists a link between the nodes $n_t$ and ${n'}_{'t}$; see also Sec.\ref{sec:eval} for details on the link prediction evaluation. In this section, we utilize a percentage $x\%$ of all the links in the graph to train the methods and the rest $95-x\%$ to test the approaches, while holding $5\%$ for validation. 

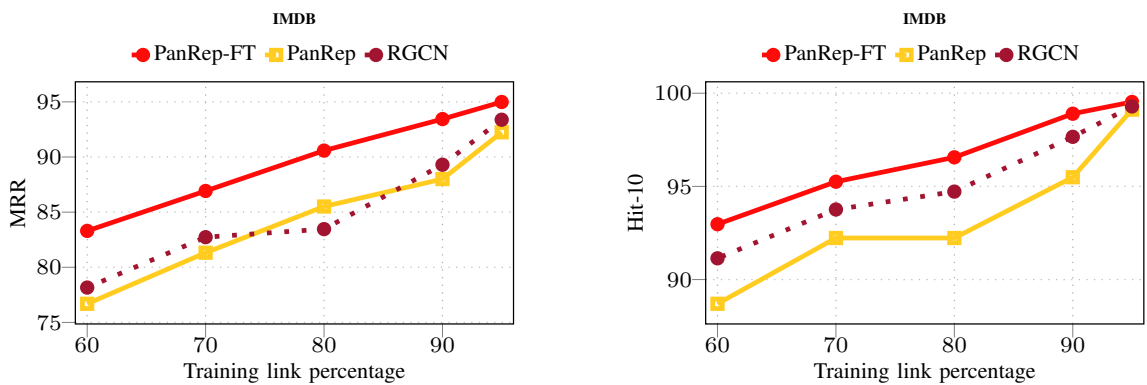
\begin{figure*}
\centering{\input{mrrnokg}~\hspace{1cm}~\input{hit10nokg}}
\caption{MRR and Hit-10 for link prediction across different percentages of testing links for IMDB.}
\label{fig:link_pred_dif_splitsimdb}
\end{figure*}

\begin{figure*}
\centering{\input{mrroag}}~\hspace{1cm}{\input{hit10oag}}
\caption{MRR and Hit-10 for link prediction across different percentages of testing links for OAG.}
\label{fig:link_pred_dif_splits}
\end{figure*}
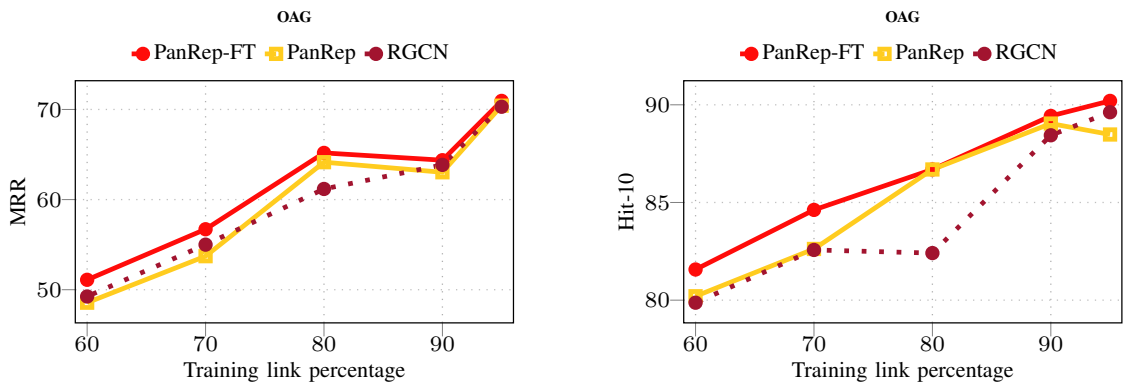

Figures~\ref{fig:link_pred_dif_splitsimdb}~and~\ref{fig:link_pred_dif_splits} report the MRR and Hit-10 scores of the baseline methods along with the PanRep and PanRep-FT methods. We report the performance of the methods for different percentages of links used for training. Observe that PanRep-FT consistently outperforms the competing methods and the performance gain increases as the percentage of training links decreases. This corroborates the advantage of pretraining GNNs for link prediction.  Note that PanRep that is trained on MRW decoder which resembles the link prediction task reports similar performance with RGCN that is trained solely in link prediction. For few training data PanRep outperforms RGCN whereas as the training data increases RGCN reduces the gap in performance and eventually outperforms PanRep. This competitive performance of PanRep highlights the use of the other decoders that regularize learning and corroborate to the success of the universal embeddings in link prediction.

\subsubsection{Drug repurposing}
Drug-repurposing aims at discovering the most effective existing drugs to treat a certain disease.
We formulate drug-repurposing as a link prediction task, where we are trying to recover the appropriate drug nodes to link to a disease node. The Drug Repurposing Knowledge Graph (DRKG)~\cite{drkg2020} collects interactions from a collection of biological databases such as Drugbank~\cite{drugbank@2017}, STRING~\cite{string@2019}, IntAct~\cite{intact} and DGIdb~\cite{dgidb@2017}.

We put forth drug-repurposing as predicting direct links in the DRKG. Here, we attempt to predict whether a drug inhibits a certain gene, which is related to the target disease.  We identify 442 genes that relate to the Covid-19 disease.  We select 8,104 FDA-approved drugs in the DRKG as candidates.
Our work focused entirely on \emph{evaluating} the ranked list of candidate drugs against a set of 32 drugs that, at the time of submission, were in Covid-19 clinical trials~\cite{gordon2020sars,zhou2020network}. Specifically, given a gene whose associated protein is a potential drug target, we rank the 8104 FDA approved drugs and select the 100 highest ranked drugs. We evaluated the performance of a method by looking at the intersection between these 100 highest ranked drugs and the 32 drugs in Covid-19 trials. In our experiments, we used 442 genes that researchers have identified as potential targets for Covid-19 and computed 442 ranked lists and associated intersections. We aggregated the results by counting the number of times each one of the 32 drugs were in these intersections.
Using  the DRKG, we compare the drug repurposing results in Covid-19 among PanRep-FT that is finetuned in link prediction and the baseline RGCN. We employ $L=1$ hidden layer with $D=600$ and train for 800 epochs both networks. For each gene node we calculate with RGCN and PanRep-FT an inhibit link score associated with every drug. Based on this score, we rank all `drug-inhibits-gene' triples per target gene.  We obtain in this way 442 ranked lists of drugs, one per gene node. Finally, to assess whether our prediction is in par with the drugs used for treatment, we check the overlap among the top 100 predicted drugs and the drugs used in clinical trials per gene.

Table~\ref{tab:predicted_drugs_gene} lists the clinical drugs included in the top-100 predicted drugs across all the genes with their corresponding number of hits for the RGCN and PanRep-FT.  Several of the widely used drugs in clinical trials appear high on the predicted list in both prediction. Furthermore, PanRep-FT reports a higher hit rate than RGCN, which corroborates the benefit of using the universal pretraining decoders. The universal representation endows PanRep with increased generalization power that allows for accurate link prediction performance when training data are extremely scarce as is the case of Covid-19. While this study, does not recommend specific drugs, it demonstrates a powerful deep learning methodology to prioritize existing drugs for further investigation, which holds the potential of accelerating therapeutic development for Covid-19.

\subsection{Ablation study}

Table~\ref{tab:node_clasabls} reports an ablation study by using different decoder subsets. The OAG graph does not have a rich motif structure and thus we did not use the motif supervision there and it is excluded from the ablation study. 
PanRep that uses all decoders obtains the best performance. The \textsc{him} and \textsc{mrw} decoders and their combination exhibit the second best performance.
Next, we compare the cluster and recover supervision decoder to the attribute generation supervision that is employed by several GNN pretraining methods~\cite{hu2020gpt}. In this experiment we employ the setting of Sec.~\ref{sec:nodecl} for node classification. The attribute generation supervision employs a 2-layer MLP network that attempts to reconstruct the original nodal attributes given the node embeddings.  In Table~\ref{tab:node_clasclustvsreconstr} we report the node classification accuracy for different number of clusters $K$ considered in the CR supervision. The CR methods outperform the attribute generation method. One reason for this is that the  GNN encoder already consideres the attributes in calculating the embeddings, and this supervision does not increase the predictive capacity of the model. On the other hand the clustering supervision allows the GNN to discover clusters of nodes with similar attributes and hence refines the learned embeddings.

\section{Conclusion}
\label{sec:cl}
This paper develops a novel framework for unsupervised learning of universal node representations on heterogenous graphs termed. PanRep supervises the GNN encoder by decoders attuned to model the clustering of local node features, structural similarity among nodes,   the local and intermediate neighborhood structure, and the mutual information among same-type nodes.  To further facilitate cases where limited labels are available we implement PanRep-FT. Experiments in node classification and link prediction corroborate the competitive performance of the learned universal node representations compared to unsupervised and semi-supervised methods. Experiments on the DRKG showcase the advantage of the universal embeddings in drug repurposing.

\bibliography{my_bibliography.bib}
\bibliographystyle{IEEEtran}
\end{document}

%% file: include.tex
\usepackage{pgfplots}
\pgfplotsset{compat=newest}
\usetikzlibrary{plotmarks}
\usetikzlibrary{positioning}
\usetikzlibrary{arrows.meta}
\usetikzlibrary{shapes,arrows}
\usetikzlibrary{fit,chains}
\usepgfplotslibrary{patchplots}
\usetikzlibrary{patterns}
\usepackage{grffile}
\usetikzlibrary{pgfplots.groupplots}
\pgfplotsset{plot coordinates/math parser=false}
\usetikzlibrary{pgfplots.units}
\pgfplotsset{plot coordinates/math parser=false}
\usepackage{grffile}

\pgfplotsset{plot coordinates/math parser=false}
\newlength\mywidth
\newlength\myheight
\newlength\mywidths
\newlength\myheights
\newlength\mywidthss
\newlength\myheightss
\definecolor{mycolor1}{rgb}{0.00000,0.44700,0.74100}%
\definecolor{mycolor2}{rgb}{0.85000,0.32500,0.9800}%
\definecolor{mycolor3}{rgb}{0.92900,0.69400,0.12500}%
\definecolor{mycolor4}{rgb}{0.89400,0.18400,0.15600}%
\definecolor{mycolor5}{rgb}{0.46600,0.67400,0.18800}%
\definecolor{mycolor6}{rgb}{0.30100,0.74500,0.93300}%

\definecolor{PanRep}{rgb}{1,0.8,0.119800}%
\definecolor{ComplEx}{rgb}{0.05000,0.32500,0.89800}%
\definecolor{PanRep-FT}{rgb}{1,0.04700,0.04100}%
\definecolor{RotatE}{rgb}{0.2,0.800,1}%
\definecolor{RGCN}{rgb}{0.63500,0.07800,0.18400}%

\definecolor{TightHann}{rgb}{0.92900,0.69400,0.12500}%
\definecolor{pDiffScater}{rgb}{0.49400,0.18400,0.55600}%
\definecolor{pMonicCubic}{rgb}{0.46600,0.67400,0.18800}%
\definecolor{pTightHann}{rgb}{0.30100,0.74500,0.93300}%

\pgfplotscreateplotcyclelist{colorlist}{%
color=AGNN,line width=2.0pt,
solid, every mark/.append style={solid, fill=AGNN}, mark=*\\%
color=GCN,line width=2.0pt,dotted, every mark/.append style={solid, fill=GCN}, mark=square*\\%
color=AGNNnf,line width=2.0pt,densely dotted, every mark/.append style={solid, fill=AGNNnf}, mark=otimes*\\%
color=Mune,line width=2.0pt,loosely dotted, every mark/.append style={solid, fill=Mune}, mark=triangle*\\%
color=mycolor5,line width=2.0pt,
dashed, every mark/.append style={solid, fill=mycolor5},mark=diamond*\\%
color=mycolor6,line width=2.0pt,loosely dashed, every mark/.append style={solid, fill=mycolor6},mark=*\\%
color=mycolor7,densely dashed, every mark/.append style={solid, fill=mycolor7},line width=2.0pt,mark=square*\\%
dashdotted, every mark/.append style={solid, fill=mycolor7},mark=otimes*\\%
dashdotdotted, every mark/.append style={solid},mark=star\\%
densely dashdotted,every mark/.append style={solid, fill=gray},mark=diamond*\\%
}
\pgfplotscreateplotcyclelist{my black white}{%
solid, every mark/.append style={solid, fill=gray}, mark=*\\%
dotted, every mark/.append style={solid, fill=gray}, mark=square*\\%
densely dotted, every mark/.append style={solid, fill=gray}, mark=otimes*\\%
loosely dotted, every mark/.append style={solid, fill=gray}, mark=triangle*\\%
dashed, every mark/.append style={solid, fill=gray},mark=diamond*\\%
loosely dashed, every mark/.append style={solid, fill=gray},mark=*\\%
densely dashed, every mark/.append style={solid, fill=gray},mark=square*\\%
dashdotted, every mark/.append style={solid, fill=gray},mark=otimes*\\%
dashdotdotted, every mark/.append style={solid},mark=star\\%
densely dashdotted,every mark/.append style={solid, fill=gray},mark=diamond*\\%
}
\relax
\usetikzlibrary{decorations.pathreplacing}
\newcommand{\legendfontsize}{\scriptsize}
\newcommand{\ticklabelfontsize}{\scriptsize}
\newcommand{\mlabelfontsize}{\scriptsize}
\definecolor{lavander}{cmyk}{0,0.48,0,0}
\definecolor{violet}{cmyk}{0.79,0.88,0,0}
\definecolor{burntorange}{cmyk}{0,0.52,1,0}
\definecolor{mygreen}{rgb}{0,1,0}%
\definecolor{myred}{rgb}{1,0,0}%

\def\halfImageWidth{8*\unit}

\newcommand{\scalelp}{1.2}

\newcommand{\mylinewidth}{1.5}
\newcommand{\markwidth}{1.5}

\tikzstyle{nnblock}=[draw,rectangle,  rounded corners,text=black,
minimum width=30pt,minimum height=60pt]
\tikzstyle{esl}=[draw,rectangle,  rounded corners,text=black,
minimum width=60pt,minimum height=30pt]
\tikzstyle{grnn}=[draw,rectangle,  rounded corners,text=black,
minimum width=30pt,minimum height=15pt]
\tikzstyle{graphsampler}=[draw,rectangle,  fill=orange,fill opacity=0.1, rounded corners,text=orange,
minimum width=30pt,minimum height=15pt]
\tikzstyle{outblock}=[draw,rectangle, color=red, rounded corners,text=black,
minimum width=10pt,minimum height=60pt]
\tikzstyle{nam}=[rectangle, fill=orange,fill opacity=0.1, rounded corners,text=orange,
minimum width=8pt,minimum height=58pt]
\tikzstyle{gam}=[fill=blue,fill opacity=0.1, rectangle,  rounded corners,text=black,
minimum width=8pt,minimum height=58pt]
\tikzstyle{fam}=[fill=green,fill opacity=0.1, rectangle,  rounded corners,text=black,
minimum width=8pt,minimum height=58pt]
\def\colorNode{node}
\def\colorEdge{edge}	 
\definecolor{edge}{cmyk}{1,0.74,0,0.77}
\definecolor{node}{cmyk}{0.99,0.66,0,0.57} 	 
\def \thisplotscale {0.75}
\pgfplotsset{
compat=1.11,
legend image code/.code={
\draw[mark repeat=2,mark phase=2]
plot coordinates {
(0cm,0cm)
(0.15cm,0cm)        
(0.3cm,0cm)         
};%
}
}
\def\myArrowWidth{1}
\definecolor{graph}{cmyk}{0.38,0.17,1,0.17} 
\def\colorGraphFilter{graph}
\def \unit {\thisplotscale cm}
\def\distNodes{0.1*\halfImageWidth}
\tikzstyle {node} = 
	[circle, 
	 draw,inner sep=0pt,
	 minimum width = 0.1*\unit,
	 minimum height = 0.1*\unit,
	 anchor = center]

%% file: motifs.tex
{\scriptsize \begin{tikzpicture}[scale=1, transform shape]
    
    
    \node at (1, 1) (o0d) {2-node motif}; 
    \node at (1.3, 0.5) (o0) {}; 
	\path (o0) 
		      node (00) [fill = \colorNode, node, draw = \colorEdge] {};
    \path (00) ++ (-1*\distNodes,0)
              node (10)[fill = \colorNode, node, draw = \colorEdge] {};
    \path (00) 
       edge [draw = \colorGraphFilter, line width = \myArrowWidth] 
       (10);
    \node at (4.5, 1) (o0d) {3-node motifs}; 
    \node at (4, 0.5) (o1) {}; 
	\path (o1) 
		      node (01) [fill = \colorNode, node, draw = \colorEdge] {};
    \path (01) ++ (-1*\distNodes,0)
              node (11)[fill = \colorNode, node, draw = \colorEdge] {};
    \path (01) 
       edge [draw = \colorGraphFilter, line width = \myArrowWidth] 
       (11);
    \path (01) ++ (-1*\distNodes,-\distNodes)
              node (21)[fill = \colorNode, node, draw = \colorEdge] {};
    \path (21) 
       edge [draw = \colorGraphFilter, line width = \myArrowWidth] 
       (11);
    
    \node at (5.5, 0.5) (o2) {}; 
	\path (o2) 
		      node (02) [fill = \colorNode, node, draw = \colorEdge] {};
    \path (02) ++ (-1*\distNodes,0)
              node (12)[fill = \colorNode, node, draw = \colorEdge] {};
    \path (02) 
       edge [draw = \colorGraphFilter, line width = \myArrowWidth] 
       (12);
    \path (02) ++ (-1*\distNodes,-\distNodes)
              node (22)[fill = \colorNode, node, draw = \colorEdge] {};
    \path (22) 
       edge [draw = \colorGraphFilter, line width = \myArrowWidth] 
       (12);
    \path (22) 
       edge [draw = \colorGraphFilter, line width = \myArrowWidth] 
       (02);
    \node at (3, -1) (o0d) {4-node motifs}; 
        \node at (0, -1.5) (o3) {}; 
	\path (o3) 
		      node (03) [fill = \colorNode, node, draw = \colorEdge] {};
    \path (03) ++ (-1*\distNodes,0)
              node (13)[fill = \colorNode, node, draw = \colorEdge] {};
    \path (03) 
       edge [draw = \colorGraphFilter, line width = \myArrowWidth] 
       (13);
    \path (03) ++ (-1*\distNodes,-\distNodes)
              node (23)[fill = \colorNode, node, draw = \colorEdge] {};
    \path (03) ++ (0,-\distNodes)
              node (33)[fill = \colorNode, node, draw = \colorEdge] {};
    \path (23) 
       edge [draw = \colorGraphFilter, line width = \myArrowWidth] 
       (13);
    \path (23) 
       edge [draw = \colorGraphFilter, line width = \myArrowWidth] 
       (33);
        \node at (1.5, -1.5) (o4) {}; 
	\path (o4) 
		      node (04) [fill = \colorNode, node, draw = \colorEdge] {};
    \path (04) ++ (-1*\distNodes,0)
              node (14)[fill = \colorNode, node, draw = \colorEdge] {};
    \path (04) 
       edge [draw = \colorGraphFilter, line width = \myArrowWidth] 
       (14);
    \path (04) ++ (-1*\distNodes,-\distNodes)
              node (24)[fill = \colorNode, node, draw = \colorEdge] {};
    \path (04) ++ (0,-\distNodes)
              node (34)[fill = \colorNode, node, draw = \colorEdge] {};
    \path (24) 
       edge [draw = \colorGraphFilter, line width = \myArrowWidth] 
       (14);
    \path (24) 
       edge [draw = \colorGraphFilter, line width = \myArrowWidth] 
       (34);
    \path (04) 
       edge [draw = \colorGraphFilter, line width = \myArrowWidth] 
       (34);
            \node at (3, -1.5) (o5) {}; 
	\path (o5) 
		      node (05) [fill = \colorNode, node, draw = \colorEdge] {};
    \path (05) ++ (-1*\distNodes,0)
              node (15)[fill = \colorNode, node, draw = \colorEdge] {};
    \path (05) 
       edge [draw = \colorGraphFilter, line width = \myArrowWidth] 
       (15);
    \path (05) ++ (-1*\distNodes,-\distNodes)
              node (25)[fill = \colorNode, node, draw = \colorEdge] {};
    \path (05) ++ (0,-\distNodes)
              node (35)[fill = \colorNode, node, draw = \colorEdge] {};
    \path (25) 
       edge [draw = \colorGraphFilter, line width = \myArrowWidth] 
       (15);
    \path (25) 
       edge [draw = \colorGraphFilter, line width = \myArrowWidth] 
       (35);
    \path (05) 
       edge [draw = \colorGraphFilter, line width = \myArrowWidth] 
       (35);
        \path (15) 
       edge [draw = \colorGraphFilter, line width = \myArrowWidth] 
       (35);
            \node at (4.5, -1.5) (o6) {}; 
	\path (o6) 
		      node (06) [fill = \colorNode, node, draw = \colorEdge] {};
    \path (06) ++ (-1*\distNodes,0)
              node (16)[fill = \colorNode, node, draw = \colorEdge] {};
    \path (06) 
       edge [draw = \colorGraphFilter, line width = \myArrowWidth] 
       (16);
    \path (06) ++ (-1*\distNodes,-\distNodes)
              node (26)[fill = \colorNode, node, draw = \colorEdge] {};
    \path (06) ++ (0,-\distNodes)
              node (36)[fill = \colorNode, node, draw = \colorEdge] {};
    \path (26) 
       edge [draw = \colorGraphFilter, line width = \myArrowWidth] 
       (16);
    \path (26) 
       edge [draw = \colorGraphFilter, line width = \myArrowWidth] 
       (36);
    \path (06) 
       edge [draw = \colorGraphFilter, line width = \myArrowWidth] 
       (36);
        \path (16) 
       edge [draw = \colorGraphFilter, line width = \myArrowWidth] 
       (36);
                \path (26) 
       edge [draw = \colorGraphFilter, line width = \myArrowWidth] 
       (06);
                    \node at (6, -1.5) (o7) {}; 
	\path (o7) 
		      node (07) [fill = \colorNode, node, draw = \colorEdge] {};
    \path (07) ++ (-1*\distNodes,0)
              node (17)[fill = \colorNode, node, draw = \colorEdge] {};
    \path (07) 
       edge [draw = \colorGraphFilter, line width = \myArrowWidth] 
       (17);
    \path (07) ++ (-1*\distNodes,-\distNodes)
              node (27)[fill = \colorNode, node, draw = \colorEdge] {};
    \path (07) ++ (0,-\distNodes)
              node (37)[fill = \colorNode, node, draw = \colorEdge] {};
    \path (27) 
       edge [draw = \colorGraphFilter, line width = \myArrowWidth] 
       (37);
    \path (07) 
       edge [draw = \colorGraphFilter, line width = \myArrowWidth] 
       (37);
        \path (17) 
       edge [draw = \colorGraphFilter, line width = \myArrowWidth] 
       (37);
                       \node at (7.5, -1.5) (o8) {}; 
	\path (o8) 
		      node (08) [fill = \colorNode, node, draw = \colorEdge] {};
    \path (08) ++ (-1*\distNodes,0)
              node (18)[fill = \colorNode, node, draw = \colorEdge] {};
    \path (08) ++ (-1*\distNodes,-\distNodes)
              node (28)[fill = \colorNode, node, draw = \colorEdge] {};
    \path (08) ++ (0,-\distNodes)
              node (38)[fill = \colorNode, node, draw = \colorEdge] {};
    \path (28) 
       edge [draw = \colorGraphFilter, line width = \myArrowWidth] 
       (38);
    \path (08) 
       edge [draw = \colorGraphFilter, line width = \myArrowWidth] 
       (38);
        \path (18) 
       edge [draw = \colorGraphFilter, line width = \myArrowWidth] 
       (38);
\end{tikzpicture}} 

%% file: mrrnokg.tex
\begin{tikzpicture}[scale=\scalelp]
\pgfplotsset{
compat=1.11,
legend image code/.code={
\draw[mark repeat=2,mark phase=2]
plot coordinates {
(0cm,0cm)
(0.125cm,0cm)        
(0.125cm,0cm)         
};%
}
}
\begin{axis}[width=0.956\mywidthss,
height=0.987\myheightss,
at={(0\mywidth,0\myheight)},
title style={at={(0.5,1.115)},font=\tiny},
title={\textbf{IMDB}},
legend style={draw=white!80.0!black},
tick align=outside,
tick pos=left,scale only axis,
x grid style={white!69.01960784313725!black},
xlabel={Training link percentage},
xmin=59, xmax=96,
xmajorgrids,
ymajorgrids,ticklabel style={font=\tiny},
grid style={dotted},
ylabel={MRR},
ticklabel style={font=\tiny, inner sep=0pt,outer sep=0pt},
legend columns=3,label style={font=\tiny},
legend style={
	at={(.1,1.015)}, 
	anchor=south west, legend cell align=left, align=left, draw=none
,font=\legendfontsize},ticklabel style={font=\ticklabelfontsize},label style={font=\mlabelfontsize}]
\addplot [line width=\mylinewidth,PanRep-FT, mark=*, mark size=\markwidth,  mark options={solid}]
table [row sep=\\]{%
95 94.99 \\
90 93.44\\
80  90.58\\		
70 86.92\\
60 83.30\\
};
\addlegendentry{PanRep-FT}

\addplot [line width=\mylinewidth,PanRep, mark=square, mark size=\markwidth, mark options={solid}]
table [row sep=\\]{%
95 92.24\\
90 88.00\\
80  85.51\\		
70 81.32\\
60 76.69\\
};
\addlegendentry{PanRep}
\addplot [line width=\mylinewidth,RGCN, mark=*, mark size=\markwidth, loosely dotted , mark options={solid}]
table [row sep=\\]{%
95 93.38\\
90 89.30\\
80  83.46\\		
70 82.73\\
60 78.16\\
};
\addlegendentry{RGCN}

\end{axis}

\end{tikzpicture}

%% file: hit10nokg.tex
\begin{tikzpicture}[scale=\scalelp]
\pgfplotsset{
compat=1.11,
legend image code/.code={
\draw[mark repeat=2,mark phase=2]
plot coordinates {
(0cm,0cm)
(0.125cm,0cm)        
(0.125cm,0cm)         
};%
}
}
\begin{axis}[width=0.956\mywidthss,
height=0.987\myheightss,
at={(0\mywidth,0\myheight)},
title style={at={(0.5,1.115)},font=\tiny},
title={\textbf{IMDB}},
legend style={draw=white!80.0!black},
tick align=outside,
tick pos=left,scale only axis,
x grid style={white!69.01960784313725!black},
xlabel={Training link percentage},
xmin=59, xmax=96,
xmajorgrids,
ymajorgrids,ticklabel style={font=\tiny},
grid style={dotted},
ylabel={Hit-10},
ticklabel style={font=\tiny, inner sep=0pt,outer sep=0pt},
legend columns=3,label style={font=\tiny},
legend style={
	at={(0.1,1.015)}, 
	anchor=south west, legend cell align=left, align=left, draw=none
,font=\legendfontsize},ticklabel style={font=\ticklabelfontsize},label style={font=\mlabelfontsize}]
\addplot [line width=\mylinewidth,PanRep-FT, mark=*, mark size=\markwidth,  mark options={solid}]
table [row sep=\\]{%
95 99.53 \\
90 98.9\\
80  96.56\\		
70 95.25\\
60 92.97\\
};

\addlegendentry{PanRep-FT}

\addplot [line width=\mylinewidth,PanRep, mark=square, mark size=\markwidth, mark options={solid}]
table [row sep=\\]{%
95 99.12\\
90 95.49\\
80 92.23\\	
70 92.23\\
60 88.70\\
};

\addlegendentry{PanRep}
\addplot [line width=\mylinewidth,RGCN, mark=*, mark size=\markwidth, loosely dotted , mark options={solid}]
table [row sep=\\]{%
95 99.29\\
90 97.66\\
80  94.72\\	
70 93.76\\
60 91.14\\
};
\addlegendentry{RGCN}

\end{axis}

\end{tikzpicture}

%% file: mrroag.tex
\begin{tikzpicture}[scale=\scalelp]
\pgfplotsset{
compat=1.11,
legend image code/.code={
\draw[mark repeat=2,mark phase=2]
plot coordinates {
(0cm,0cm)
(0.125cm,0cm)        
(0.125cm,0cm)         
};%
}
}
\begin{axis}[width=0.956\mywidthss,
height=0.987\myheightss,
at={(0\mywidth,0\myheight)},
title style={at={(0.5,1.115)},font=\tiny},
title={\textbf{OAG}},
legend style={draw=white!80.0!black},
tick align=outside,
tick pos=left,scale only axis,
x grid style={white!69.01960784313725!black},
xlabel={Training link percentage},
xmin=59, xmax=96,
xmajorgrids,
ymajorgrids,ticklabel style={font=\tiny},
grid style={dotted},
ylabel={MRR},
ticklabel style={font=\tiny, inner sep=0pt,outer sep=0pt},
legend columns=3,label style={font=\tiny},
legend style={
	at={(0.1,1.015)}, 
	anchor=south west, legend cell align=left, align=left, draw=none
,font=\legendfontsize},ticklabel style={font=\ticklabelfontsize},label style={font=\mlabelfontsize}]
\addplot [line width=\mylinewidth,PanRep-FT, mark=*, mark size=\markwidth,  mark options={solid}]
table [row sep=\\]{%
95 70.95 \\
90 64.37\\
80  65.18\\		
70 56.71\\
60 51.10 \\
};
\addlegendentry{PanRep-FT}

\addplot [line width=\mylinewidth,PanRep, mark=square, mark size=\markwidth, mark options={solid}]
table [row sep=\\]{%
95 70.42\\
90 63.03\\
80 64.15\\	
70 53.72 \\
60 48.56\\
};

\addlegendentry{PanRep}
\addplot [line width=\mylinewidth,RGCN, mark=*, mark size=\markwidth, loosely dotted , mark options={solid}]
table [row sep=\\]{%
95 70.30\\
90 63.85\\
80  61.18\\	
70 55.00\\
60 49.25\\
};
\addlegendentry{RGCN}

\end{axis}

\end{tikzpicture}

%% file: hit10oag.tex
\begin{tikzpicture}[scale=\scalelp]
\pgfplotsset{
compat=1.11,
legend image code/.code={
\draw[mark repeat=2,mark phase=2]
plot coordinates {
(0cm,0cm)
(0.125cm,0cm)        
(0.125cm,0cm)         
};%
}
}
\begin{axis}[width=0.956\mywidthss,
height=0.987\myheightss,
at={(0\mywidth,0\myheight)},
title style={at={(0.5,1.115)},font=\tiny},
title={\textbf{OAG}},
legend style={draw=white!80.0!black},
tick align=outside,
tick pos=left,scale only axis,
x grid style={white!69.01960784313725!black},
xlabel={Training link percentage},
xmin=59, xmax=96,
xmajorgrids,
ymajorgrids,ticklabel style={font=\tiny},
grid style={dotted},
ylabel={Hit-10},
ticklabel style={font=\tiny, inner sep=0pt,outer sep=0pt},
legend columns=3,label style={font=\tiny},
legend style={
	at={(0.1,1.015)}, 
	anchor=south west, legend cell align=left, align=left, draw=none
,font=\legendfontsize},ticklabel style={font=\ticklabelfontsize},label style={font=\mlabelfontsize}]
\addplot [line width=\mylinewidth,PanRep-FT, mark=*, mark size=\markwidth,  mark options={solid}]
table [row sep=\\]{%
95 90.20 \\
90 89.43\\
80  86.69\\		
70 84.62\\
60 81.57\\
};
\addlegendentry{PanRep-FT}

\addplot [line width=\mylinewidth,PanRep, mark=square, mark size=\markwidth, mark options={solid}]
table [row sep=\\]{%
95 88.48\\
90  89.04\\	
80 86.69\\
70 82.62\\
60 80.20\\
};
\addlegendentry{PanRep}
\addplot [line width=\mylinewidth,RGCN, mark=*, mark size=\markwidth, loosely dotted , mark options={solid}]
table [row sep=\\]{%
95 89.62\\
90 88.44\\
80  82.41\\		
70 82.57\\
60 79.87\\
};

\addlegendentry{RGCN}

\end{axis}

\end{tikzpicture}